\documentclass{article}

\PassOptionsToPackage{numbers,compress}{natbib}
\usepackage[main,final]{neurips_2026}

\usepackage[utf8]{inputenc}
\usepackage[T1]{fontenc}
\usepackage{hyperref}
\hypersetup{hidelinks}
\usepackage{url}
\usepackage{booktabs}
\usepackage{amsmath,amssymb,amsfonts}
\usepackage{graphicx}
\usepackage{subcaption}
\usepackage{microtype}
\usepackage[capitalise]{cleveref}
\usepackage{placeins}
\usepackage{xcolor}         

\title{Cool Graphs: Active Property Search Towards Quantum Nano-Refrigerators}

\author{%
  Sofia Blyufer \\
  Schulich Faculty of Chemistry \\
  Technion -- Israel Institute of Technology \\
  Haifa, Israel \\
  \texttt{sofiablyufer@campus.technion.ac.il} \\
  \And
  Amir Kleiner\thanks{Corresponding authors.} \\
  Department of Molecular Chemistry \\ and Materials Science \\
  Weizmann Institute of Science \\
  Rehovot, Israel \\
 \texttt{amir.kleiner@weizmann.ac.il} \\
  \And
  Uri Peskin$^{*}$ \\
  Schulich Faculty of Chemistry and \\
  Helen Diller Quantum Center \\
  Technion -- Israel Institute of Technology \\
  Haifa, Israel \\
  \texttt{uri@technion.ac.il} \\
}

\begin{document}
\maketitle

\begin{abstract}
Nanodevices that absorb energy from their thermal environment can act as quantum refrigerators. Finding suitable configurations, however, is a needle-in-a-haystack problem: the many-body state space scales exponentially with the size of the device, where each candidate configuration requires an expensive quantum-chemical computation. 
We introduce a search framework that couples a \emph{Graph Variational Autoencoder} (GVAE) with an uncertainty-driven \emph{Active Learning} (AL) loop to efficiently identify energy-absorbing configurations in electronic junctions with a 2D "working material" to be optimized.
The GVAE maps physical parameters - site energies, coupling matrices, temperature - into a smooth latent space that distinguishes cooling from dissipative parameter regimes, achieving a classification discriminability of $0.920 \pm 0.004$ on held-out configurations and $0.895 \pm 0.010$ on unseen extrapolation topologies.
An AL pipeline based on Monte Carlo dropout then targets the decision boundary, iteratively selecting the candidates where the model is least certain for simulation.
Over 32 AL cycles, the uncertainty-driven strategy discovers energy absorbing systems at rates consistently exceeding the random baseline across all tested topologies (where only 1.9-8.4\% of configurations exhibit the desired property), providing a data-efficient pathway for navigating a combinatorically large design space. 
The learned latent space organizes configurations along physically interpretable axes, that correlate with underlying microscopic model parameters. 
This hints at the existence of "cooling coordinates" - low-dimensional,
physically interpretable quantities separating cooling from non-cooling regimes. Our framework, thus, paves a new way towards effective cooling of nanoscale electronic information processing devices, to reduce the environmental footprint of high performance computing.
\end{abstract}

\section{Introduction}
\label{sec:intro}

As electronic devices shrink to the nanoscale, classical heat-dissipation strategies reach fundamental limits~\cite{pop2010energy,cahill2003nanoscale,myers2025future}.
In this regime, energy transport is governed by the dynamics of open quantum systems, whose discrete electronic states couple to fermionic reservoirs (electrodes) and bosonic (e.g. phonon) heat baths~\cite{dubi2011colloquium,thoss2018perspective}.
Theoretical studies have shown that certain nanoscale junctions show unique transport features that differ significantly from those of macroscopic devices. Of particular interest are energy pumps that can extract heat from their environment opposing an external temperature gradient~\cite{dubi2011colloquium,thoss2018perspective}, suggesting the plausibility of nanoscale refrigeration based on quantum transport
principles~\cite{cleuren2012cooling,levy2012quantum}. Quantum absorption refrigeration has been demonstrated experimentally in a trapped-ion system~\cite{maslennikov2019quantum}. Most propositions for quantum refrigerators rely on a few level systems, whose embedding in realistic devices remains a remarkable challenge. 

Here we focus on nanoscale conductors undergoing charge and energy transport. Our aim is to explore and identify system configurations capable of absorbing energy from their (bosonic) environment in nonequilibrium conditions, as potential candidates for cooling devices. Discovering such configuration in complex quantum conductors is, however, a formidable search problem.
Calculating the steady state energy currents into (or out of) a quantum system of $N$ single particle states (orbitals) involves the solution of a master equation in the electronic Fock space, whose dimension is $2^N$. Consequently, the computational effort scales exponentially with the system size~\cite{iv2024emergence}; measured timings for our oracle are given in \cref{app:oracle_cost}.
Worse, system configurations involving energy absorption turn out to be rare. Considering 2D systems of different sizes and shapes, we found that under uniform random sampling of the physical parameter space, only 1.2-9.2\% of configurations exhibit positive bosonic energy flux, with the higher rates corresponding to the simpler and smaller systems, falling to 1.2\% for the larger 2D grids. Exhaustive or random search is therefore inefficient~\cite{cleuren2012cooling,levy2012quantum,dubi2011colloquium,thoss2018perspective}, where physics-based guiding principles are limited to the smallest model systems. 

Active learning has emerged as a key strategy for reducing the cost of computational screening in materials discovery~\cite{settles2009active,lookman2019active,kusne2020fly}.
Graph neural networks provide natural representations for molecular and crystalline systems~\cite{gilmer2017neural,bronstein2021geometric}, and variational graph autoencoders~\cite{kipf2016variational,simonovsky2018graphvae} offer smooth latent spaces suitable for optimization and uncertainty quantification. Equivariant architectures such as SchNet, DimeNet, PaiNN, and NequIP~\cite{schutt2017schnet,gasteiger2020directional,schutt2021equivariant,batzner2022e3} have become standard for predicting quantum-chemical properties from atomic geometries. Our graphs instead encode the parameters of a model Hamiltonian (on-site energies and couplings) rather than atomic coordinates, so their symmetry constraints do not directly apply.
\citet{gomez2018automatic} demonstrated that property-conditioned latent spaces enable targeted molecular search, and at larger scale \citet{merchant2023scaling} used deep learning to accelerate inorganic crystal discovery. Closer to our application, \citet{erdman2023model} used model-free reinforcement learning to optimize the driving cycles of quantum heat engines and refrigerators: we address a complementary problem, searching over static device configurations for those having refrigeration potential, namely that can absorb energy from their boson surroundings during operation.
Generative flow networks~\cite{bengio2021flow,jain2023gflownets} offer another route to rare-candidate discovery by sampling diverse high-reward objects, and are a natural complement to the generative use of our decoder.

In this work, we frame the discovery task of energy-absorbing nano-systems as a \emph{guided search} problem in the space of the candidate system Hamiltonians and propose a framework that combines three ingredients:
\textbf{(i)}~A physics-based oracle that solves the steady-state master equation for any candidate junction. \textbf{(ii)}~A Graph Variational Autoencoder (GVAE) that learns a compressed latent space separating energy absorbing from energy dissipating regimes, where the variational formulation is chosen deliberately so that the latent space is semantic and continuous with the underlying physical parameters, making it potentially interpretable and therefore a source of physical insight.
\textbf{(iii)}~An Active Learning (AL) loop that uses Monte Carlo (MC) dropout uncertainty to select the most informative candidates for simulation, concentrating expensive oracle calls where they matter most.
To the best of our knowledge, this is the first work to combine graph-based latent representations with active learning for the discovery of refrigeration regimes in quantum conductors.

\section{The Physical System and Motivation to the Search Problem}
\label{sec:phys_model}

We consider a 2D system of $n = n_\text{row}\times n_\text{col}$ electronic sites arranged in a grid, connected to two fermionic electrodes (left/right) and $n$ local bosonic heat baths.
The single-particle Hamiltonian reads
$[\hat{h}_\text{sys}]_{ij} = \alpha_i \delta_{ij} + \beta_0\, e^{-\lambda_M (r_{ij} - 1)}(1-\delta_{ij})$,
where $\alpha_i$ is the on-site energy, $\beta_0$ the inter-site coupling parameter, $\lambda_M$ a spatial decay constant, and $r_{ij}$ the Euclidean distance between sites.
Each site ($i$) couples to the left/right electrode ($J$) with effective strength $\eta^{i,J} = \eta_0^J e^{-2\lambda_E (r_{i,J} - 1)}$, and to the bosonic baths via a normalized coupling matrix~$W$ (see Appendix ~\ref{app:model_system}).
To simplify this proof of concept electron-electron interactions are neglected, and only rectangular grids were considered; the graph formulation itself supports electron-electron interactions and non-rectangular geometries.
The system and all couplings are illustrated in figure~\ref{fig:sys2graph}(a).

To account for the reduced quantum dynamics and energy fluxes we invoke the Born-Markov and secular approximations ~\cite{breuer2002theory,peskin2023quantum}, applicable in the limit of non-degenerate system Hamiltonian and weak coupling between the system and the external reservoirs. This is an intentional choice as it preserves the transport features controlled by the device's internal structure, whereas strong coupling hybridizes system and reservoir states and washes those features out.
Fixed and equal reservoir temperatures are assumed following the same reasoning, since internal reservoir relaxation is faster than charge and energy exchange with the system, where  the reservoirs remain nearly at the ambient equilibrium temperature even during transport.

Working in the many-body Fock space of dimension $2^N$, we construct rate matrices for fermionic ($K^{\text{fer}}_{L/R}$) and bosonic ($K^{\text{bos}}$) induced transitions between the many-body eigenstates of the system Hamiltonian (see Appendix ~\ref{app:transport}). Each rate depends on the Fermi-Dirac or Bose-Einstein distributions, respectively~\cite{breuer2002theory,peskin2023quantum}.
The vector of steady-state populations of many-body eigenstates $P^{ss}$ satisfies $K_\text{tot}\,P^{ss} = 0$ with $K_\text{tot} = K_L^{\text{fer}} + K_R^{\text{fer}} + K^{\text{bos}}$, and the bosonic energy flux is $F_E = \mathbf{E}^\top K^{\text{bos}} P^{ss}$,
where $\mathbf{E}$ is the vector of many-body state energies (see Appendix ~\ref{app:observables}).
A positive $F_E$ indicates energy absorption from the bosonic environment, which is a necessary condition for a \emph{nano-refrigerator} operation. 
\newline
Analysis of a minimal two-site model (see Appendix~\ref{app:two_level}) reveals that positive  fluxes can be obtained when the steady state population ratio between a high-energy orbital and a low energy one is kept at a sub-Boltzmann level (considering the reservoirs' temperature), resulting in a constant energy current from the boson reservoirs. Such sub-Boltzmann orbital populations can be obtained under a potential bias, where the low/high energy orbital is coupled to the electrode with the high/low chemical potential, which implies that the system orbitals must be localized in different spatial regions.  Yet, the  efficiency of the energy absorption also depends on the local overlap between the two orbitals, resulting in a narrow parameter regime, in which the two conflicting demands are simultaneously met. For larger, realistic systems, formulation of the necessary conditions for energy absorption in the vast parameter space becomes intractable, motivating a data-driven approach.

\section{Methods}
\label{sec:methods}

Each data point is generated by sampling physical parameters from the ranges in \Cref{tab:params}, constructing and solving the many-body master equation (see Appendices~\ref{app:model_system} - ~\ref{app:observables} for the full derivation and computational procedure), and recording the resulting $F_E$. 
To avoid ambiguities arising from near-zero flux values, we define the binary label as positive only for systems with substantial energy absorption, setting the threshold at $F_E > 10^5$~eV/s; this value lies well below typical positive-flux magnitudes ($10^7$-$10^{10}$~eV/s) while clearly excluding the dissipative regime (see Appendix~\ref{app:flux_hist} for the full flux distribution). The threshold also mitigates the dependence on the oracle's precision: the sensitivity near $F_E \approx 0$ comes from the steady-state null vector, and since that vector is positive and normalized, problematic relative errors arise mainly when its entries become nearly uniform, which does not typically occur in the finite-bias, finite-temperature regime considered here.

\begin{figure}[h]
    \centering
    \includegraphics[width=\linewidth]{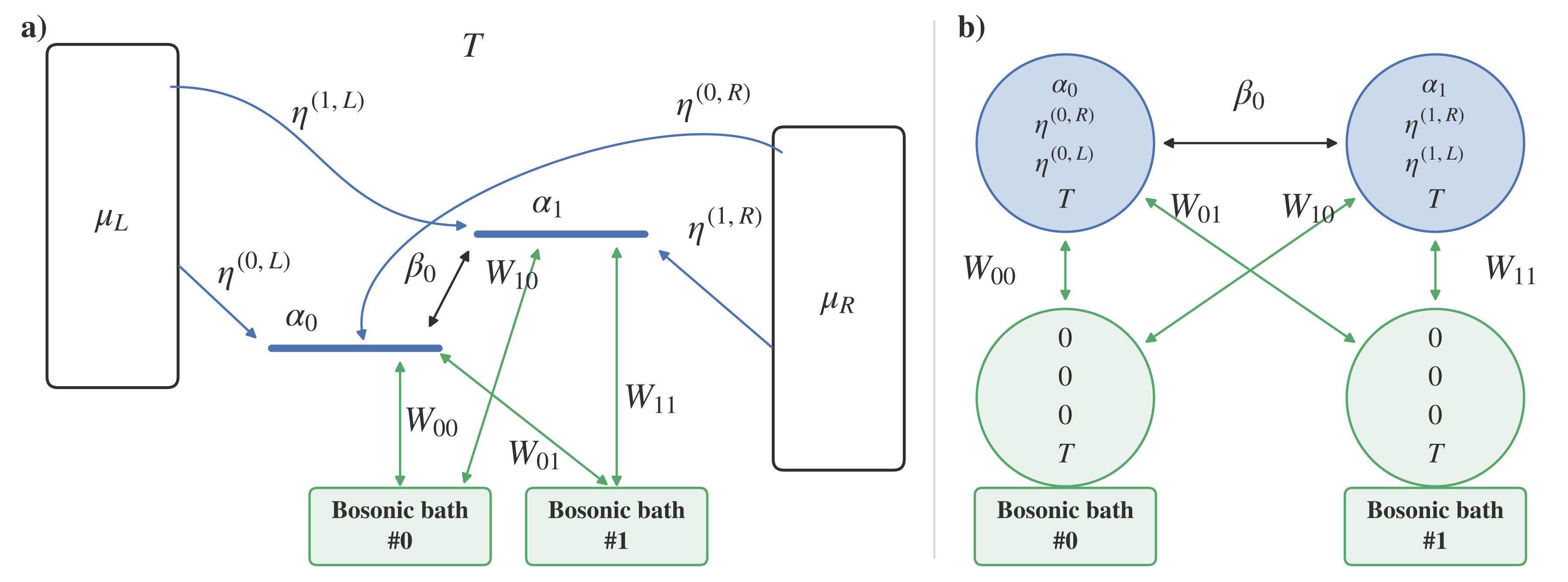}
    \caption{System-to-graph mapping. \textbf{(a)}~The physical junction with two fermionic electrodes (with chemical potentials $\mu_L$  and $\mu_R$), electronic sites ($\alpha_0$, $\alpha_1$), inter-site coupling ($\beta_0$), bosonic bath couplings ($W_{ij}$), and electrode couplings ($\eta^{(i,J)}$) at temperature $T$. \textbf{(b)}~The corresponding attributed graph: site nodes carry features $[\alpha_i, T, \eta^{(i,L)}, \eta^{(i,R)}]$, bath nodes carry $[0, 0, 0, T]$, and edges carry coupling strengths from $\hat{h}_\text{sys}$ and $W$.}
    \label{fig:sys2graph}
\end{figure}

The physical system is mapped to an undirected graph $\mathcal{G} = (\mathcal{V}, \mathcal{E})$ following~\cite{battaglia2018relational}: molecular sites and bosonic baths become nodes with features $\mathbf{x}_i = [\alpha_i,\, T,\, \eta^{(i,L)},\, \eta^{(i,R)}]^\top$ (bath nodes set non-$T$ features to zero), and edges carry weights from $\hat{h}_\text{sys}$ (site-site) or $W$ (site-bath), as illustrated in \Cref{fig:sys2graph}b.
To handle the wide dynamic range of $F_E$, all features are normalized via a signed-log1p transform followed by z-scoring.

\begin{table}[h]
\centering
\caption{Sampling ranges for data generation. $T$, $\beta_0$, $\lambda_M$,
$\lambda_E$, $\eta_0^{L/R}$ and $\mu_{L/R}$ are as defined in
\Cref{sec:phys_model}. The on-site energies are generated by sampling a base
value $\alpha_\text{base}$ and a spacing $\Delta\alpha$, setting
$\alpha_i = \alpha_\text{base} + \pi(i)\Delta\alpha$, where $\pi$ randomly
permutes the sites so that the energy ordering is uncorrelated with position.}
\label{tab:params}
\small
\begin{tabular}{@{}ll@{}}
\toprule
\textbf{Parameter} & \textbf{Range} \\ \midrule
Grid sizes (train) & $1{\times}2$, $1{\times}3$, $1{\times}4$, $2{\times}2$, $2{\times}3$, $2{\times}4$, $3{\times}2$, $3{\times}3$ \\
Grid sizes (extrap.) & $1{\times}5$,\; $3{\times}4$,\; $4{\times}2$ \\
$T$ & $[150,\;500]$~K \\
$\beta_0$ & $[-0.15,\;-0.01]$~eV \\
$\lambda_M,\,\lambda_E$ & $[5,\;1000]$ \\
$\eta_0^{L/R}$ & $10^x$,\; $x\!\in\![\log_{10}0.01,\;\log_{10}0.2]$~eV \\
$W_\text{scale}$ & $10^x$,\; $x\!\in\![\log_{10}0.001,\;\log_{10}0.2]$~eV \; (\Cref{app:bath}) \\
$\tilde{W}$ pattern & 10 bath-connectivity patterns (\Cref{app:bath}) \\
$\alpha_\text{base}$ & $[-0.1,\;1.2]$~eV \\
$\Delta\alpha$ & $[-2|\beta_0|,\;2|\beta_0|]$~eV \\
$\mu_L,\ \mu_R$ & $4$, $0$~eV \; (fixed, not sampled) \\
\bottomrule
\end{tabular}
\end{table}

\begin{figure}[h]
    \centering
    \begin{subfigure}[t]{\linewidth}
        \centering
        \includegraphics[width=\linewidth]{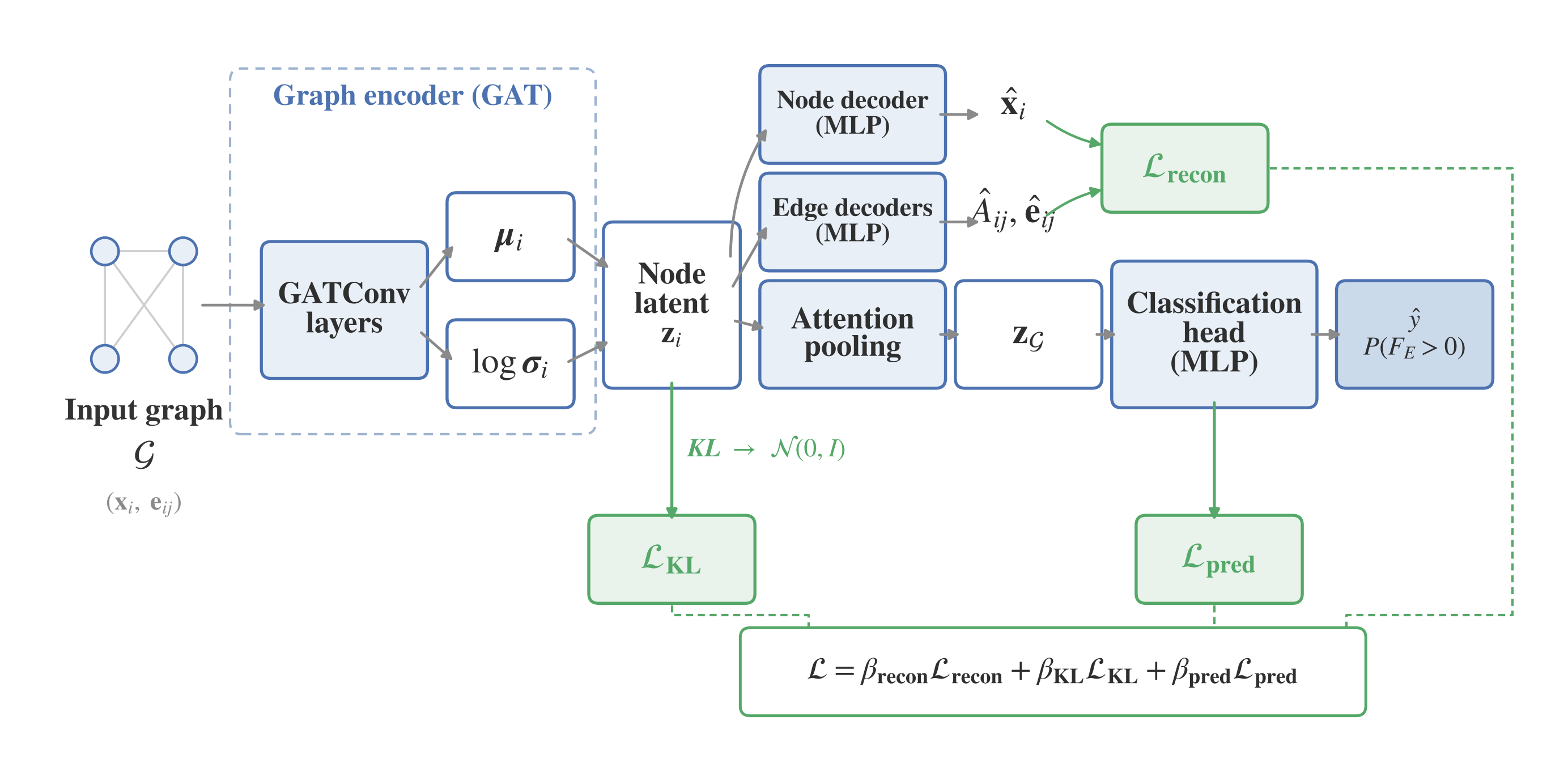}
        \caption{GVAE architecture with losses $\mathcal{L}_\text{recon}$, $\mathcal{L}_\text{KL}$, and $\mathcal{L}_\text{pred}$.}
        \label{fig:gvae_arch}
    \end{subfigure}
    \\[1em]
    \begin{subfigure}[t]{\linewidth}
        \centering
        \includegraphics[width=0.7\linewidth]{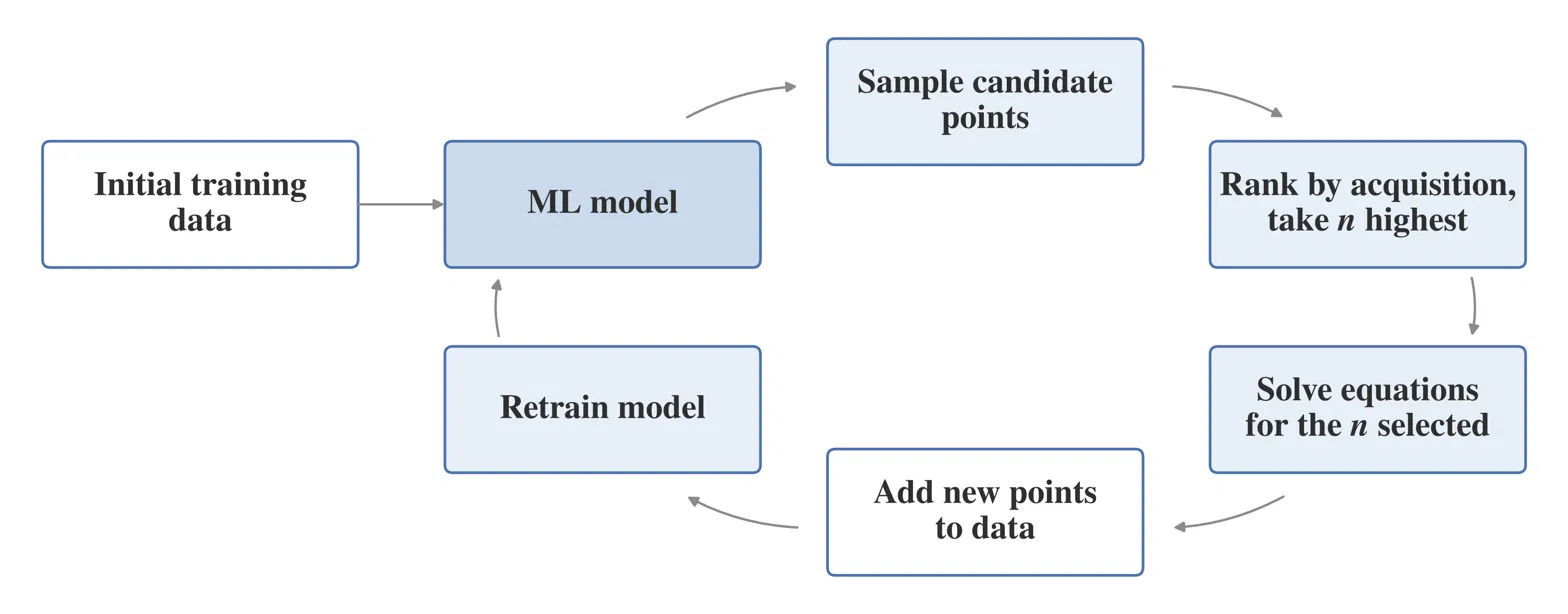}
        \caption{Active learning loop.}
        \label{fig:al_cycle}
    \end{subfigure}
    \caption{Model and search pipeline. \textbf{(a)}~The GVAE encodes graphs into a variational latent space; auxiliary decoders ($\mathcal{L}_\text{recon}$) regularize the representation while a classification head ($\mathcal{L}_\text{pred}$) learns the positive-flux decision boundary, balanced by KL regularization ($\mathcal{L}_\text{KL}$). \textbf{(b)}~The AL loop uses MC-dropout uncertainty to select maximally informative candidates for expensive master-equation simulation, iteratively enriching the training set.}
    \label{fig:framework}
\end{figure}

The role of the GVAE (\Cref{fig:framework}a) is to learn a smooth, low-dimensional landscape over which the search can operate; rather than searching directly in the high-dimensional physical parameter space, the AL pipeline queries an organized latent space where nearby points share similar transport properties. Beyond the classification task, the variational formulation is chosen to yield a structured latent space amenable to subsequent physical interpretation and uncertainty quantification, both of which are essential for the active learning loop. This approach follows a growing body of work demonstrating that property-conditioned VAE latent spaces enable targeted search in molecular~\cite{gomez2018automatic,liu2018constrained} and materials~\cite{sanchez2018inverse} design.
A multi-head Graph Attention Network (GAT)~\cite{velickovic2018graph} encoder maps node features and edge attributes to contextualized embeddings, which are projected to variational parameters $\boldsymbol{\mu}_i$ and $\log\boldsymbol{\sigma}_i$ defining $q_\phi(\mathbf{z}_i \mid \mathcal{G})$.
Auxiliary MLP decoders reconstruct node features and edge weights from the latent embeddings, serving as regularizers that encourage the latent space to retain physical information.
Attention-based pooling~\cite{li2016gated,lee2019set} assigns learnable importance weights to node embeddings to produce a fixed-size graph representation $\mathbf{z}_\mathcal{G}$, from which an MLP classification head maps to the positive-flux probability, providing the decision boundary that the AL pipeline targets.
The total loss balances reconstruction, Kullback-Leibler (KL) regularization~\cite{kingma2019introduction,higgins2017beta}, and prediction (classification):
$\mathcal{L} = \beta_\text{recon}\,\mathcal{L}_\text{recon} + \beta_\text{KL}\,\mathcal{L}_\text{KL} + \beta_\text{pred}\,\mathcal{L}_\text{pred}$.
Training proceeds in two stages: pure autoencoder pretraining ($\beta_\text{KL}{=}\beta_\text{pred}{=}0$), followed by full multi-objective training.
Hyperparameters were selected via Bayesian optimization (Optuna~\cite{optuna_2019}) maximizing validation area under the precision-recall curve (PR-AUC) on the validation dataset (see Appendix~\ref{app:hyperparams} for the full search space), yielding latent dimension $32$, hidden dimension $128$, $\beta_\text{recon}{=}2.67$, $\beta_\text{KL}{=}5.9\times10^{-3}$, $\beta_\text{pred}{=}65.8$.

To efficiently expand the training set toward rare positive-flux configurations, we employ an uncertainty-driven AL loop (\Cref{fig:framework}b).
In each cycle, a large pool of unlabeled candidate systems is sampled from the parameter distributions, and $T{=}20$ stochastic forward passes are performed through the GVAE with MC dropout~\cite{gal2016dropout} to estimate the predictive variance $\sigma^2 = \frac{1}{T}\sum_t (\hat{y}_t - \bar{y})^2$ for each candidate.
The $k$ candidates with highest $\sigma^2$ are selected, their $F_E$ values are computed via the master-equation solver, and the newly labeled points are added to the training set before retraining the GVAE.
A single active-learning run proceeds for 32 cycles; in each cycle 10{,}000 candidates are freshly sampled, the 500 with the highest acquisition score are simulated, and the model is retrained on the enlarged pool, at a median cost of 57 minutes per cycle (hardware, software stack and model size in \cref{app:resources}).
We call one such complete run of 32 cycles and 16{,}000 oracle calls a \emph{trajectory}, and repeat each acquisition strategy for three independent trajectories that differ only in the random seed.
MC dropout is inexpensive but can at times yield poorly calibrated uncertainty estimates~\cite{djupskas2026unreliable}; to mitigate this, we also run the loop with a five-member deep ensemble~\cite{lakshminarayanan2017simple} as the source of acquisition uncertainty, under the same protocol and budget.
Acquisition is purely uncertainty-driven; periodically replacing an acquisition with random selection, as an exploration safeguard, is examined as an ablation in \cref{app:ablations}. 
Predictive results are reported as mean $\pm$ sample standard deviation over five training seeds, scored on a fixed stratified split so that every model and search strategy is evaluated on the same held-out sets. Active-learning results are reported as mean $\pm$ sample standard deviation over the three trajectories of each strategy.
As a non-graph baseline we also run Gaussian-process Bayesian optimization at the same acquisition budget.
A Gaussian process requires a fixed-dimensional input, so one independent GP is fitted per topology over the continuous parameters of \Cref{tab:params} with a Mat\'ern-$5/2$ plus white-noise kernel, and the 16{,}000 calls are divided evenly across the eight training topologies.
At each step the GP scores a fresh pool of 1{,}200 candidates drawn from the same sampler the AL loop uses and the highest-scoring one is simulated, so both methods draw candidates from the same distribution, though the surrogates ranking them differ in what they have seen: the graph model starts from the 28{,}000-graph pool, while each GP begins from 12 randomly drawn points in its own topology, paid out of the same 16{,}000-call budget, and is refitted on at most 300 points at a time.
The objective is a signed-log transform of $F_E$; the kernel hyper-parameters are re-optimized every 25 steps and the posterior updated at every step. We report both expected improvement (EI) and upper confidence bound (UCB), each over three trajectories of 16{,}000 calls with different random seeds, as for the active-learning strategies.
The held-out rows of \Cref{tab:al} are produced identically for every strategy: a GVAE is retrained from scratch on the initial 28{,}000-graph pool plus that strategy's 16{,}000 acquisitions, under the protocol above, and scored on the same frozen held-out sets, so those rows differ only in which points were acquired.
\FloatBarrier

\section{Results and Discussion}
\label{sec:results}

The system was investigated in 11 distinct topologies, ranging from 1D chains of up to 5 sites to 2D grids of up to $3{\times}4$ sites.
Under uniform random sampling, positive-flux configurations are exceedingly rare, with positive hit rates ranging from 9.2\% down to 1.2\% with the hit rates of 2D grids consistently lower than
the 1D chains (\Cref{tab:scarcity}), a trend that is expected to continue for larger systems.
The training set is drawn from this natural distribution: 40{,}000 graphs across the 8 training topologies ($1{\times}2$, $1{\times}3$, $1{\times}4$, $2{\times}2$, $2{\times}3$, $2{\times}4$, $3{\times}2$, $3{\times}3$) with an overall positive rate of 4.25\%.
The remaining 3 topologies were held out entirely and form the \emph{extrapolation test set} (15{,}000 graphs); a stratified split of the 8 training topologies supplies the validation set and the \emph{interpolation test set} (6{,}000 graphs each).

\begin{table}[h]
\centering
\caption{Positive-flux hit rates under random sampling, illustrating the scarcity of the target regime.}
\label{tab:scarcity}
\small
\begin{tabular}{@{}lcc|lcc@{}}
\toprule
\textbf{Topology} & \textbf{Samples} & \textbf{Hit rate} & \textbf{Topology} & \textbf{Samples} & \textbf{Hit rate} \\ \midrule
$1{\times}4$ & 5000 & 8.38\% & $2{\times}3$ & 5000 & 3.52\% \\
$1{\times}3$ & 5000 & 6.58\% & $3{\times}3$ & 5000 & 1.86\% \\
$1{\times}2$ & 5000 & 4.08\% & $3{\times}2$ & 5000 & 2.08\% \\
$2{\times}4$ & 5000 & 3.80\% & $2{\times}2$ & 5000 & 3.66\% \\
\midrule
\multicolumn{6}{@{}l@{}}{\emph{Extrapolation topologies, absent from training}} \\
$1{\times}5$ & 6000 & 9.15\% & $4{\times}2$ & 6000 & 1.18\% \\
$3{\times}4$ & 3000 & 1.93\% & & & \\
\bottomrule
\end{tabular}
\end{table}

\begin{figure}[h]
    \centering
    \includegraphics[width=\linewidth]{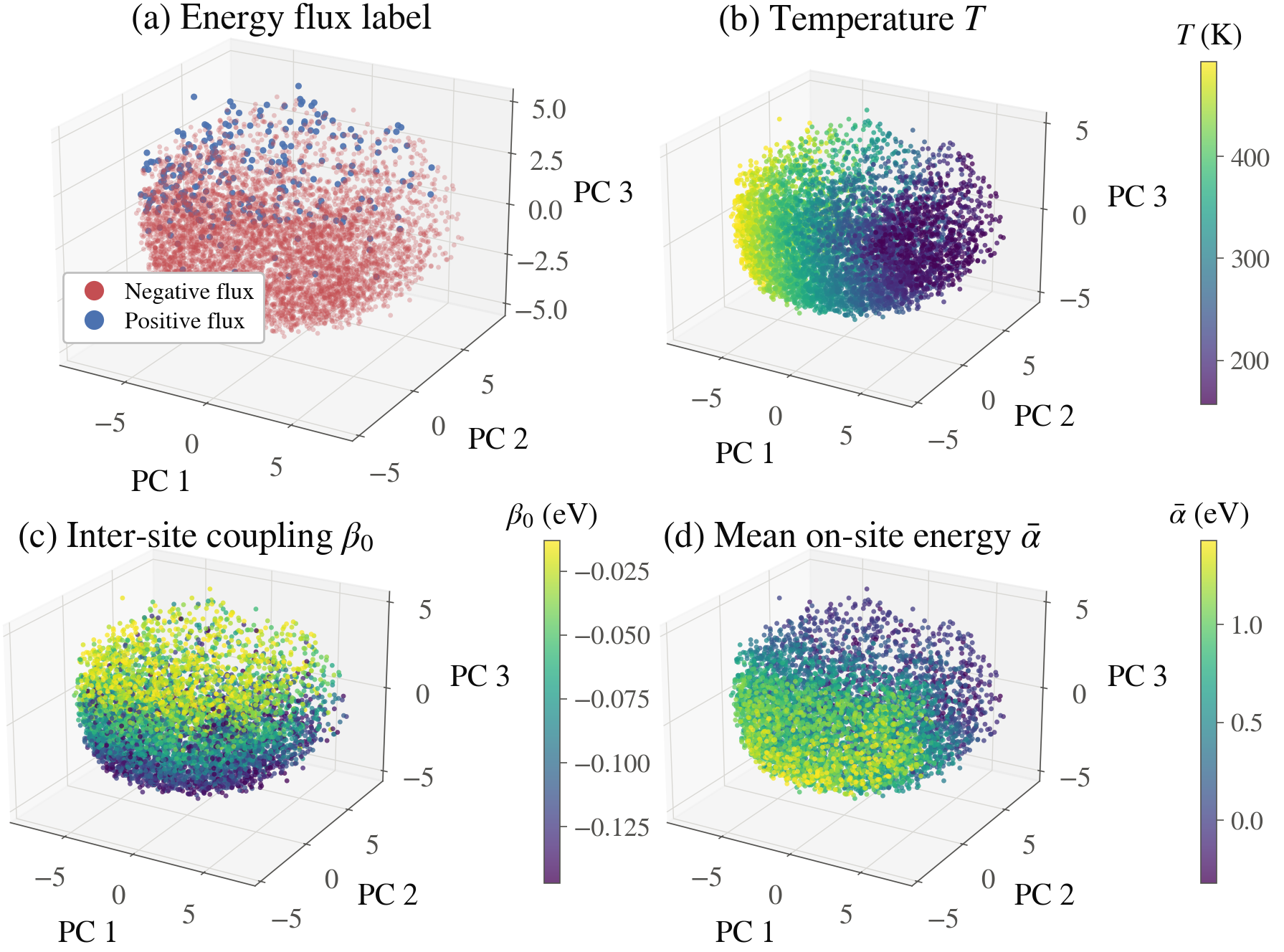}
    \caption{PCA projection of the graph-level latent space $\mathbf{z}_{\mathcal{G}}$ for one GVAE seed on the interpolation test set; the three components capture 40.1\%, 22.6\% and 9.5\% of the variance (72.2\% in total), and all three axes are principal components. \textbf{(a)}~Positive (refrigerating, blue; $n=254$) and negative (dissipative, red; $n=5746$) configurations. \textbf{(b-d)}~The same projection colored by physical parameters, none of which is supplied as a training target.}
    \label{fig:latent}
\end{figure}

The GVAE achieves a receiver operating characteristic area under the curve (ROC-AUC) of $0.920 \pm 0.004$ on the interpolation test set and $0.895 \pm 0.010$ on the extrapolation test set ($1{\times}5$, $3{\times}4$, $4{\times}2$; 15{,}000 graphs, 678 positives, 4.52\%) entirely absent from training, one of which ($3{\times}4$) reaches $N{=}12$ against a maximum of $N{=}9$ in training.
The Latent space representation of the systems was visualized by means of principal component analysis (PCA), where the first three principal components, capturing 72.2\% of the total variance, were chosen for visualization purposes. 
\Cref{fig:latent}a shows the positive (negative) flux labels of the systems by color-coding them blue (red).

Positive-flux systems occupy a restricted parameter regime of, among others, small $|\beta_0|$ and narrow $\alpha$ ranges, consistent with the two-level localization intuition. Training is confined to systems of at most $N{=}9$ orbitals, where the oracle remains affordable (\cref{app:oracle_cost}); the extrapolation set reaches $N{=}12$. 

Although the model did not train to predict the system's physical parameters, most of them are, nevertheless, linearly decodable from the latent space, and are ordered mostly monotonically along a leading principal component.
For example consider the temperature, with linear relation of $R^2 = 0.995 \pm 0.002$ and primary component rank correlation of $0.984 \pm 0.014$ over five seeds.\cref{app:ablations}). 
The latent space interpertability of the inter-site couplings and mean on-site energy as well as temperature are illustrated in figure~\ref{fig:latent}.

Considering that trained networks have been shown to be readable for the physics they have learned~\citep{amey2021neural},
our present results suggest that the cooling regime could be characterized by interpretable, dimensionless quantities extracted from the latent space.


Our central results are summarized in \Cref{tab:al}. Over 32 cycles and 16{,}000 oracle calls, uncertainty-driven acquisition discovers $4477 \pm 105$ positive-flux configurations against $670 \pm 14$ for random acquisition at the same budget (mean $\pm$ s.d. over three trajectories), so each expensive master-equation evaluation is $6.68\times$ more likely to return a refrigerating configuration.
Enrichment holds on every topology and is largest where positives are rarest: $8.35\times$ on $3{\times}3$ (1.86\% base rate) against $4.20\times$ on the $1{\times}4$ chain (8.38\%).
For comparison, a model was also trained adding 16{,}000 points acquired at random, giving a second pool of 44{,}000 graphs at the sampler's 4.2\% positive rate.
The discovered points also improve the decision boundary: at the same final pool size, test ROC-AUC reaches $0.946 \pm 0.011$ against $0.932 \pm 0.002$ for random acquisition and extrapolation ROC-AUC $0.934 \pm 0.006$ against $0.911 \pm 0.006$. 
Against Gaussian-process Bayesian optimization at the same acquisition budget, uncertainty-driven acquisition finds $1.46$-$1.53\times$ more positives and is markedly more reproducible: the Gaussian-process trajectories span 2547-3695 positives under UCB and 1855-4007 under EI, against 4374-4584 for MC dropout.
The held-out rows separate discovery from learning. Both Gaussian-process pools carry over four times the random strategy's positives, raising the training pool's positive rate from 4.2\% to 9.4\% and 9.7\%, yet neither trains a better classifier than random acquisition: test ROC-AUC is $0.924 \pm 0.012$ and $0.923 \pm 0.012$ against $0.932 \pm 0.002$, differences within the spread across trajectories.
Were the number of positives the operative variable, the Gaussian-process strategies would be expected to recover much of the gap; they recover none of it.
This suggests that positives acquired near the decision boundary are more informative to the model than an equal or larger number acquired by objective-driven search---a hypothesis this experiment motivates but does not establish, since testing it requires pools matched on positive count, which we leave to future work.

Three structural asymmetries qualify the comparison.
The budgets are matched on acquisition only: every active-learning strategy, random included, starts from the same 28{,}000-graph labelled pool, which the Gaussian process does not use, so by the end the graph surrogate has seen 28{,}000 more oracle evaluations than the GP.
A Gaussian process also requires a fixed-dimensional input, so one is fitted per topology: the baseline shares no information between geometries and cannot be applied to an unseen one at all, whereas the graph surrogate transfers. It divides its budget evenly across topologies while the active-learning loop reallocates, though \cref{app:topology_enrichment} shows this does not account for the gap --- the loop spends more of its budget on $1{\times}4$ and still loses there, and less on the 2D grids and still wins.
Conversely, the baseline examines far more candidates per acquired point than the loop does, 1{,}200 against 20, so the candidate budget does not handicap it.
The baseline is a standard Gaussian-process implementation run cold: seeded from twelve random points per topology, refitted on at most 300 of its accumulated evaluations, and driven by acquisitions that maximize the flux rather than seek its threshold. Warm-starting it from the existing labelled pool, replacing the exact GP with a sparse approximation, or adopting a level-set acquisition matched to the counting metric would each strengthen it, and none is tested here.

\begin{table}[h]
\centering
\caption{Search results at a matched budget of 16{,}000 oracle calls, three independent trajectories per strategy. The held-out rows are obtained by retraining a GVAE from scratch on the shared 28{,}000-graph initial pool plus that strategy's 16{,}000 acquisitions, giving a 44{,}000-graph pool in every case, so they isolate which points were acquired rather than how many; the rows above them concern the search itself, which for GP-BO uses no initial pool. Enrichment is the ratio of positives found to the random strategy at the same budget. The two GP-BO columns are the same Gaussian-process surrogate under different acquisition functions; it divides its budget evenly across the eight topologies, because a Gaussian process requires a fixed-dimensional input space, whereas the active-learning loop reallocates (\cref{app:topology_enrichment}).}
\label{tab:al}
\small
\setlength{\tabcolsep}{4pt}
\begin{tabular}{@{}lccccc@{}}
\toprule
 & \textbf{Random} & \textbf{MC-dropout} & \textbf{GP-BO (EI)} & \textbf{GP-BO (UCB)} & \textbf{Deep ensemble} \\ \midrule
Positives found              & $670 \pm 14$      & $4477 \pm 105$       & $2923 \pm 1076$      & $3073 \pm 580$       & $4925 \pm 47$ \\
Oracle calls per positive    & $23.88 \pm 0.49$  & $3.58 \pm 0.08$      & $6.04 \pm 2.36$      & $5.33 \pm 0.98$      & $3.25 \pm 0.03$ \\
Enrichment over random       & $1.00$            & $6.68\times$         & $4.36\times$         & $4.58\times$         & $7.35\times$ \\
\quad range across topologies & ---              & $4.20$--$8.35\times$ & $1.35$--$6.15\times$ & $2.21$--$6.54\times$ & $4.65$--$9.92\times$ \\
Final pool positive rate     & $4.23\%$          & $12.88\%$            & $9.35\%$             & $9.69\%$             & $13.90\%$ \\
Interpolation ROC-AUC        & $0.932 \pm 0.002$ & $0.946 \pm 0.011$    & $0.924 \pm 0.012$    & $0.923 \pm 0.012$    & $0.942 \pm 0.002$ \\
Extrapolation ROC-AUC        & $0.911 \pm 0.006$ & $0.934 \pm 0.006$    & $0.909 \pm 0.014$    & $0.897 \pm 0.010$    & $0.904 \pm 0.014$ \\
\bottomrule
\end{tabular}
\end{table}


\FloatBarrier
\section{Conclusions}
\label{sec:conclusions}

We have presented a guided-search framework for discovering nanoscale electronic devices that absorb thermal energy from their surroundings. 
Systems that support positive bosonic energy flux are physically plausible but combinatorically rare.
By learning a latent landscape that separates cooling from dissipative regimes and targeting the decision boundary with uncertainty-driven active learning, the pipeline discovers positive-flux configurations at rates far exceeding random exploration while generalizing to unseen topologies.
Acquisition uses MC-dropout variance at the classifier head, so it is uncertainty near the decision boundary that drives the enrichment, as in the discriminative baselines of \cref{app:baselines}; the latent space instead organizes the search space and provides physical interpretability, and we do not claim it causes the enrichment. 
Using the decoder generatively is beyond the scope of the present work, and so is the experimental realizability. 
Although quantum absorption refrigeration has been demonstrated experimentally for a three mode trapped-ion system~\cite{maslennikov2019quantum}, the complexity of a many-body transport problem with competing quantum resonances, electronic correlations, vibronic coupling and thermalization does not allow quantitative predictions, and motivates surrogate-guided search with more realistic models. 
A most promising result of our study is the semantic latent space, which hints at the existence of "cooling coordinates" - low-dimensional, physically interpretable quantities separating cooling from non-cooling regimes. Extracting the physical information underlying these coordinates from the latent vectors would provide both a closed-form design criterion and deeper physical understanding, which is beyond the scope of this work. 

\paragraph{Code availability:}
The code for the oracle, the GVAE, the active-learning loop and the baselines, together with the scripts that generate all datasets used here, is publicly available at \url{https://github.com/sblyufer/cool-graphs} and archived on Zenodo~\citep{blyufer2026coolgraphscode}.

\begin{ack}
This research was supported by the Technion Nevet Quantum grant, the Israel Science Foundation (Grant No.~2256/22) and the United States-Israel Binational Science Foundation (Grant No.~2020327). The authors declare no competing interests.
\end{ack}

\bibliographystyle{unsrtnat}
\bibliography{references}

\clearpage
\appendix
\crefalias{section}{appendix}
\crefalias{subsection}{subappendix}
\crefalias{subsubsection}{subsubappendix}
\makeatletter
\renewcommand{\@seccntformat}[1]{\ifcsname prefix@#1\endcsname\csname prefix@#1\endcsname\fi\csname the#1\endcsname\quad}
\newcommand{\prefix@section}{Appendix~}
\makeatother

\section{The Model System}
\label{app:model_system}

The 2D system modeled as a junction consists of $n = n_\text{row} \cdot n_\text{col}$ electronic sites arranged in a grid-like structure. Each site is coupled to $n$ bosonic baths via a connectivity matrix $\hat{W}$ (where a naive coupling of each site to a single local bath corresponds, e.g., to coupling to local phonon modes) and to two fermionic electrodes.

The open system of two electrodes, $n$ sites, and $n$ baths is described by the total Hamiltonian:
\begin{equation}
    \hat{H}_\text{total} = \hat{H}_\text{S} + \hat{H}_\text{R} + \hat{H}_\text{RS},
\end{equation}
where $\hat{H}_\text{S}$ is the system Hamiltonian, $\hat{H}_\text{R}$ is the Hamiltonian of the bosonic and fermionic reservoirs, and $\hat{H}_\text{RS}$ is the interaction Hamiltonian. The system is weakly coupled to the environment, such that each reservoir is in its own quasi-equilibrium and characterized by one or more of the following parameters:
$\mu_{L/R}$, the chemical potential of the left/right electrode;
$T$, the temperature of all reservoirs (taken to be the same for simplicity);
$W_{ij}$, the coupling of the $i$th site to the $j$th bath (normalized so that $\sum_{i,j} W_{ij} = 1$ to ensure the overall coupling does not increase with system size);
and $\eta_0^{L/R}$, the coupling strength to the left/right electrode, with effective coupling $\eta^{i,J} = \eta_0^J e^{-2\lambda_E (r_{i,J} - 1)}$, where $\lambda_E$ is a decay constant and $r_{i,J}$ is the minimal distance between the $i$th site and electrode $J$.

The single-particle orbital energies $\{\varepsilon_n\}_{n=0}^{N-1}$ and orbitals $\{v_n\}_{n=0}^{N-1}$ are obtained by diagonalizing the single-particle Hamiltonian $\hat{h}_\text{sys}$:
\begin{equation}
    [\hat{h}_\text{sys}]_{ij} =
    \begin{cases}
        \alpha_i & \text{if } i = j, \\
        \beta_0\, e^{-\lambda_M (r_{ij} - 1)} & \text{if } i \neq j,
    \end{cases}
\end{equation}
where $\alpha_i$ is the on-site energy, $\beta_0$ is the inter-site coupling parameter, $\lambda_M$ is the material decay parameter, and $r_{ij}$ is the Euclidean distance between sites. In this formulation, $v_n^i$ is the projection of the $n$th orbital on the $i$th local site.

To transition from the single-particle to the many-body description, we employ the formalism of second quantization. The fermionic annihilation and creation operators for local sites are $d_i$ and $d_i^\dagger$, and for single-particle orbitals are $a_n$ and $a_n^\dagger$. The local annihilation operator can be expanded as:
\begin{equation}
    d_i = \sum_{n=0}^{N-1} v_n^i\, a_n.
    \label{eq:si_expansion_d}
\end{equation}
The many-body system Hamiltonian is:
\begin{equation}
    \hat{H}_\text{S} = \sum_{n=0}^{N-1} \varepsilon_n\, a_n^\dagger a_n,
\end{equation}
where electron--electron interactions are excluded for simplicity. For $N$ orbitals, the many-body Fock space has dimension $M = 2^N$, and each state $|m\rangle$ specifies the occupation of every orbital. For example, $|m\rangle = |1100\rangle$ indicates that orbitals $\varepsilon_0$ and $\varepsilon_1$ are occupied while $\varepsilon_2$ and $\varepsilon_3$ are empty.

The action of the creation operator $a_n^\dagger$ on a many-body state $|m\rangle = |o_1, o_2, \dots, o_N\rangle$ is:
\begin{equation}
    a_n^\dagger |o_1, \dots, o_n, \dots, o_N\rangle =
    \begin{cases}
        (-1)^{\sum_{i=1}^{n-1} o_i}\, |o_1, \dots, 1_n, \dots, o_N\rangle & \text{if } o_n = 0, \\
        0 & \text{if } o_n = 1,
    \end{cases}
\end{equation}
where the phase factor accounts for the antisymmetry required by the Pauli exclusion principle~\cite{peskin2023quantum}.

\section{Transport}
\label{app:transport}

The master equation for the eigenstate populations reads~\cite{peskin2023quantum}:
\begin{equation}
    \frac{\partial}{\partial t} P_m(t) = \sum_{m'} \left( \sum_{J \in \{L,R\}} \left[k_J^{(\text{fer})}\right]_{m,m'} + \sum_{n_b=1}^{N} \left[k_{n_b}^{(\text{bos})}\right]_{m,m'} \right) P_{m'}(t),
\end{equation}
where $P_m(t)$ is the population of the $m$th many-body state, $N$ is the number of baths, and $J \in \{L,R\}$ is the electrode index.

The rate constants correspond to absorption or emission of a single electron or phonon:
\begin{align}
    \left[k_J^{(\text{fer})}\right]_{m,m'} &= (1 - \delta_{m,m'})(\Gamma_{m,m'}^{J;e} + \Gamma_{m,m'}^{J;h}) - \delta_{m,m'} \sum_{m'' \neq m} (\Gamma_{m'',m}^{J;e} + \Gamma_{m'',m}^{J;h}), \label{eq:si_fermion_rates} \\
    \left[k_{n_b}^{(\text{bos})}\right]_{m,m'} &= (1 - \delta_{m,m'})(\Gamma_{m,m'}^{n_b;\text{em}} + \Gamma_{m,m'}^{n_b;\text{ab}}) - \delta_{m,m'} \sum_{m'' \neq m} (\Gamma_{m'',m}^{n_b;\text{em}} + \Gamma_{m'',m}^{n_b;\text{ab}}). \label{eq:si_boson_rates}
\end{align}

\subsection{Fermionic Transition Rates}

The fermionic transition rate for electrode $J$ is:
\begin{equation}
    \Gamma_{m,m'}^{J;h/e} = \frac{1}{\hbar} \left| \sum_{i=0}^{N-1} \eta^{i,J} \langle m | d_i | m' \rangle \right|^2 f_J^{(h/e)}(E_{m'} - E_m),
\end{equation}
where $E_m = \sum_n \varepsilon_n\, o_{n}^{(m)}$ is the many-body state energy.
Using the expansion from \Cref{eq:si_expansion_d}, this becomes $\left| \sum_{i=0}^{N-1} \sum_{n=0}^{N-1} \eta^{i,J} v_n^i \langle m | a_n | m' \rangle \right|^2$.
This expression is non-zero only when $|m\rangle$ and $|m'\rangle$ differ in exactly one orbital: for holes, $o_n^{(m)} = 0$ and $o_n^{(m')} = 1$; for electrons, $o_n^{(m)} = 1$ and $o_n^{(m')} = 0$.
For example, in a two-orbital system, the non-zero electron transitions are:
\begin{equation}
    (|m\rangle, |m'\rangle) \in \{(|00\rangle, |01\rangle),\; (|00\rangle, |10\rangle),\; (|01\rangle, |11\rangle),\; (|10\rangle, |11\rangle)\}.
\end{equation}

The electrode spectral density $J_J(\varepsilon)$ is treated within the wide-band approximation and absorbed as a constant scaling factor into the coupling parameters $\eta^{i,J}$.
The Fermi--Dirac distributions are:
\begin{equation}
    f_J^e(\varepsilon) = \frac{1}{\exp\!\left(\frac{\varepsilon - \mu_J}{k_B T}\right) + 1}, \qquad f_J^h(\varepsilon) = 1 - f_J^e(\varepsilon),
\end{equation}
where $\mu_J$ is the chemical potential of electrode $J$.

\subsection{Bosonic Transition Rates}

The bosonic transition rate for bath $n_b$ is:
\begin{equation}
    \Gamma_{m,m'}^{n_b;\text{em/ab}} = \frac{1}{\hbar} \left| \langle m | \sum_{i=0}^{N-1} W_{i,n_b}\, d_i^\dagger d_i | m' \rangle \right|^2 g_{n_b}^{(\text{em/ab})}(E_{m'} - E_m),
\end{equation}
where the spectral density of the bosonic baths is assumed constant and absorbed into $W$.
Using the orbital expansion (\Cref{eq:si_expansion_d}), $\langle m | \sum_i W_{i,n_b} d_i^\dagger d_i | m' \rangle = \langle m | \sum_i W_{i,n_b} \sum_{n,n'} v_n^i v_{n'}^i\, a_n^\dagger a_{n'} | m' \rangle$.
This is non-zero only when $|m\rangle$ and $|m'\rangle$ differ in exactly two orbitals---one switching from occupied to empty and the other from empty to occupied.
For emission, $n > n'$; for absorption, $n' > n$.
For a two-orbital system, the only non-zero bosonic transition is $(|m\rangle, |m'\rangle) = (|01\rangle, |10\rangle)$.

The Bose--Einstein distributions are:
\begin{equation}
    g_{n_b}^{\text{ab}}(\varepsilon) = \frac{1}{\exp\!\left(\frac{\varepsilon}{k_B T}\right) - 1}, \qquad g_{n_b}^{\text{em}}(\varepsilon) = 1 + g_{n_b}^{\text{ab}}(\varepsilon).
\end{equation}

\section{Observables}
\label{app:observables}

The total rate matrix is:
\begin{equation}
    K_\text{tot} = K_L^{\text{fer}} + K_R^{\text{fer}} + K^{\text{bos}},
\end{equation}
where $K^{\text{bos}} = \sum_{n_b} k_{n_b}^{(\text{bos})}$.

The steady-state population vector $P^{ss}$ satisfies $K_\text{tot}\, P^{ss} = 0$ with $\sum_m P_m^{ss} = 1$, where $P_m^{ss}$ is the steady-state population of many-body state $|m\rangle$.
In practice, $P^{ss}$ is obtained by computing the null space of $K_\text{tot}$ via singular value decomposition and normalizing.

The energy density matrix is defined as $[\rho_S]_{m,m'} = \delta_{m,m'} P_m$, and the energy flux from the environment into the system is:
\begin{equation}
    \frac{\partial}{\partial t}\,\text{tr}(\rho_S \hat{H}_S) = \text{tr}\!\left[\left(\frac{\partial}{\partial t} \rho_S\right) \hat{H}_S\right].
\end{equation}
Since the contributions from all reservoirs are additive in the weak-coupling limit ($\frac{\partial}{\partial t} \rho_S = K_\text{tot}\, \rho_S$), the energy flux from the bosonic reservoirs alone is:
\begin{equation}
    F_E = \text{tr}\!\left[K^{\text{bos}} \rho_S\, \hat{H}_S\right] = \mathbf{E}^\top K^{\text{bos}} P^{ss},
    \label{eq:si_flux}
\end{equation}
where $\mathbf{E}$ is the vector of many-body state energies.
The units of $F_E$ are eV/s, determined by $E$ in eV, $K$ in s$^{-1}$ (with $\hbar$ in eV$\cdot$s), and $P$ being dimensionless.
A positive $F_E$ indicates net energy absorption from the bosonic environment.

\section{Positive Energy Flux in a Two-Level Model System}
\label{app:two_level}

In a two-site system ($N = 2$), the single-particle Hamiltonian is:
\begin{equation}
    \hat{h}_\text{sys} = \begin{pmatrix} \alpha_0 & \beta_0 \\ \beta_0 & \alpha_1 \end{pmatrix},
\end{equation}
yielding orbitals $v_0 = \begin{pmatrix} v_0^0 \\ v_0^1 \end{pmatrix}$, $v_1 = \begin{pmatrix} v_1^0 \\ v_1^1 \end{pmatrix}$ and orbital energies $\varepsilon_0, \varepsilon_1$.
The many-body states are $|00\rangle$, $|01\rangle$, $|10\rangle$, $|11\rangle$, with state-energies in ascending order.
For simplicity, $\lambda_E = 1000$ so that the left and right electrodes are only connected to the left and right sites respectively.
Each bath is connected solely to a single site, giving $W = \text{scale} \cdot \tilde{I}$, where $\tilde{I}$ is the normalized identity matrix.

The fermionic rate matrix for the right electrode is:
\begin{equation}
K_R^{\text{fer}} =
{\scriptsize
\begin{pmatrix}
-(\Gamma_{01}^{R,e} + \Gamma_{10}^{R,e}) & \Gamma_{00}^{R,h} & \Gamma_{00}^{R,h} & 0 \\
\Gamma_{01}^{R,e} & -(\Gamma_{00}^{R,h} + \Gamma_{11}^{R,e}) & 0 & \Gamma_{01}^{R,h} \\
\Gamma_{10}^{R,e} & 0 & -(\Gamma_{00}^{R,h} + \Gamma_{11}^{R,e}) & \Gamma_{10}^{R,h} \\
0 & \Gamma_{11}^{R,e} & \Gamma_{11}^{R,e} & -(\Gamma_{01}^{R,h} + \Gamma_{10}^{R,h})
\end{pmatrix}
}
\end{equation}
where, for example, $\Gamma_{|10\rangle \to |11\rangle}^{R,e} = \frac{1}{\hbar} |\eta^{1,R} v_1^1|^2 f_R^e(\varepsilon_1)$.

The bosonic rate matrix is:
\begin{equation}
K^{\text{bos}} =
{\scriptsize
\begin{pmatrix}
0 & 0 & 0 & 0 \\
0 & -(\Gamma_{01 \to 10}^{\text{ab},0} + \Gamma_{01 \to 10}^{\text{ab},1}) & \Gamma_{10 \to 01}^{\text{em},0} + \Gamma_{10 \to 01}^{\text{em},1} & 0 \\
0 & \Gamma_{01 \to 10}^{\text{ab},0} + \Gamma_{01 \to 10}^{\text{ab},1} & -(\Gamma_{10 \to 01}^{\text{em},0} + \Gamma_{10 \to 01}^{\text{em},1}) & 0 \\
0 & 0 & 0 & 0
\end{pmatrix}
}
\end{equation}
where, for example, $\Gamma_{|10\rangle \to |01\rangle}^{0,\text{em}} = \frac{1}{\hbar} |W_{i,0} v_0^0 v_1^0|^2 g_0^{\text{em}}(\varepsilon_1 - \varepsilon_0)$.

Finding $P^{ss} = (P_{00},\, P_{01},\, P_{10},\, P_{11})^\top$ from $K_\text{tot} P^{ss} = 0$ yields the energy flux $F_E = \mathbf{E}^\top K^{\text{bos}} P^{ss}$.

\begin{figure}[h]
    \centering
    \includegraphics[width=0.8\linewidth]{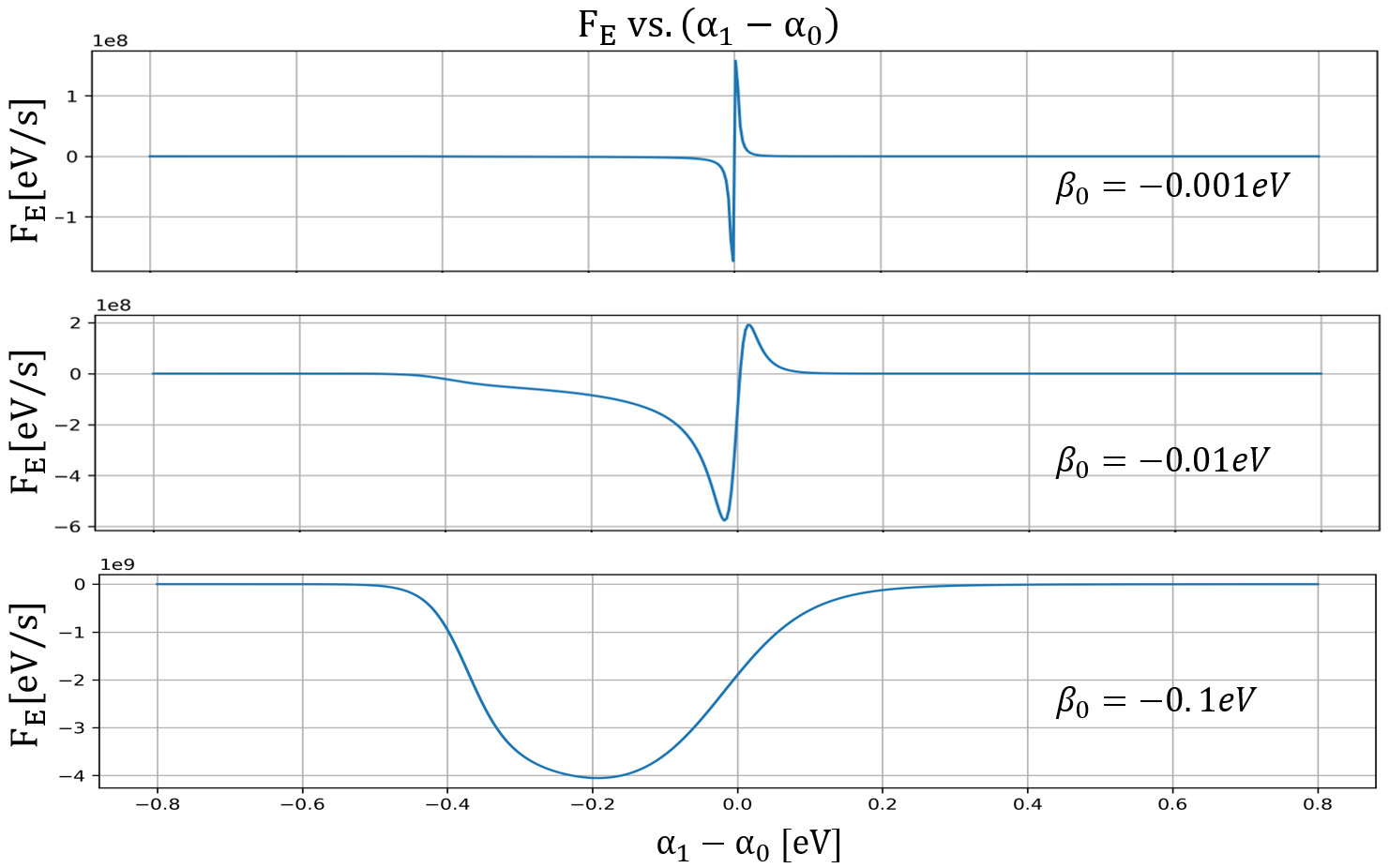}
    \caption{Energy flux $F_E$ for a two-level system at varying $\beta_0$. Parameters: $\alpha_0 = 0.4$~eV, $\alpha_1 \in [-0.4, 1.2]$~eV, $W = 0.015 \cdot \tilde{I}$, $T = 298$~K, $\lambda_M = \lambda_E = 1000$, $\eta_0^{L/R} = 0.015$~eV, $\mu_L = 4$~eV, $\mu_R = 0$. Top: $\beta_0 = -0.01$~eV; middle: $\beta_0 = -0.001$~eV; bottom: $\beta_0 = -0.1$~eV.}
    \label{fig:2level_results}
\end{figure}

\Cref{fig:2level_results} shows $F_E$ computed for varying site energy differences ($\alpha_1 - \alpha_0$) at different $\beta_0$ values.
When $|\beta_0|$ is small relative to the energy offset, the orbitals are highly localized at the contact sites.
In this regime, the orbital coupled to the electrode with higher chemical potential is preferentially occupied, while the other is mainly empty.
When the occupied orbital has the lower energy, the steady-state population ratio between the high- and low-energy orbitals is kept below the Boltzmann level, resulting in net energy absorption from the bosonic reservoirs ($F_E > 0$)---the nano-refrigerator regime.
Conversely, when the occupied orbital has higher energy, energy is dissipated ($F_E < 0$).
If $|\beta_0|$ is too large, the orbitals delocalize, the occupation differences vanish, and the system becomes purely dissipative.

\FloatBarrier
\section{Bath Connectivity}
\label{app:bath}

The formalism of \Cref{app:model_system} admits an arbitrary site--bath coupling
matrix; the patterns described here are a sampling choice, not a restriction of
the model. The coupling is drawn as $W = W_\text{scale}\,\tilde{W}$, where
$W_\text{scale}$ sets the overall strength and the pattern $\tilde{W}$ is drawn
per configuration from ten options. Each is normalized to
$\sum_{ij}\tilde{W}_{ij} = 1$, so the total coupling is fixed at
$W_\text{scale}$ independently of system size and the patterns differ only in
how it is distributed: those confined to fewer sites couple those sites more
strongly. Three are diagonal, coupling each site to its own bath---e.g.\ local
phonon modes---with equal, linearly decreasing or random strengths (0.35 in
total); six couple baths only to the leftmost or rightmost column, in some
variants extending into the adjacent column or sharing a bath between
neighbours, leaving the remaining sites uncoupled from the bosonic environment
(0.60); and one couples every site to every bath equally, which suppresses the
bosonic rates almost entirely and is therefore sampled least (0.05).

\FloatBarrier
\section{Flux Distribution and Threshold Justification}
\label{app:flux_hist}

\begin{figure}[h]
    \centering
    \includegraphics[width=0.7\linewidth]{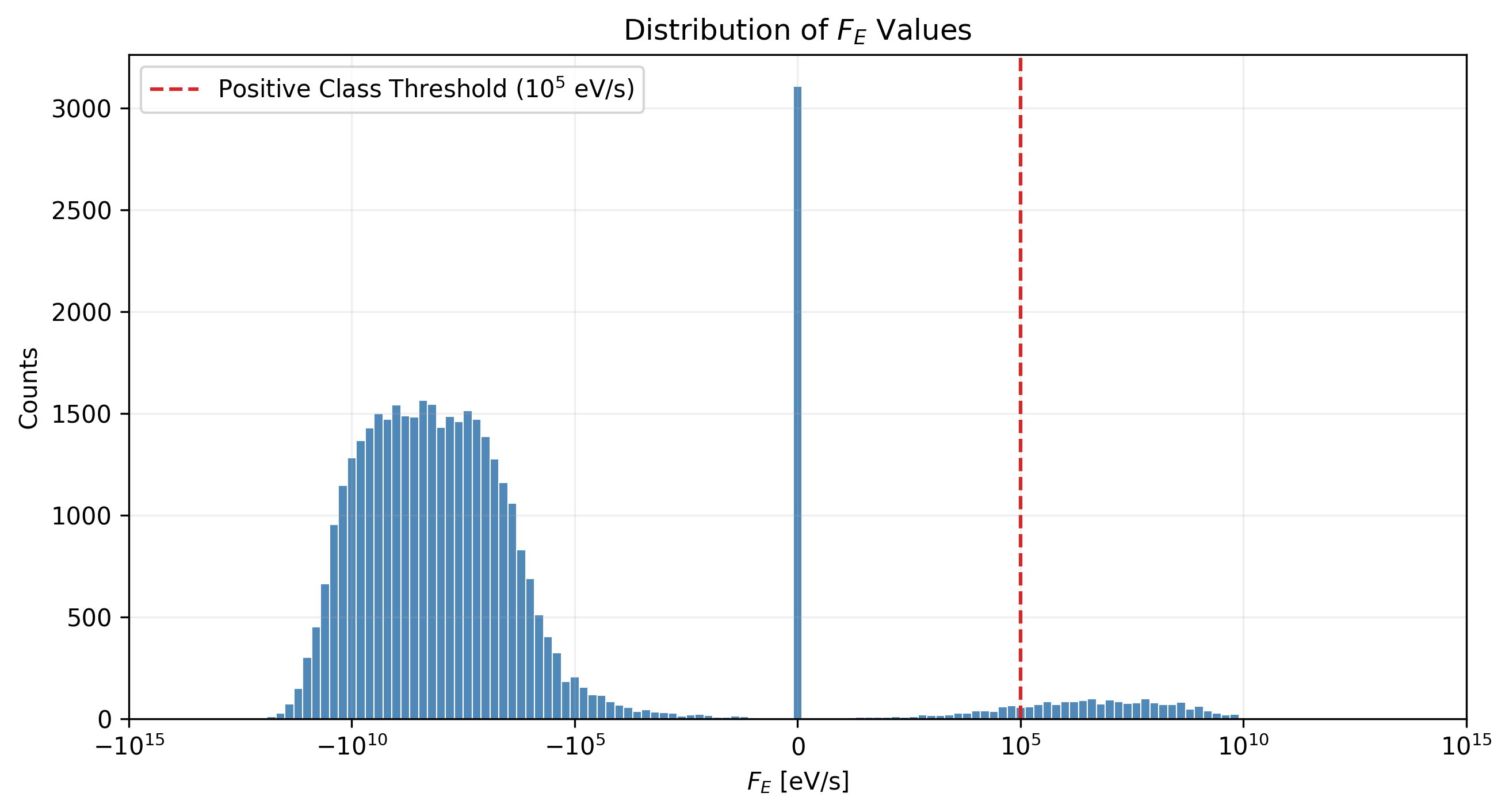}
    \caption{Distribution of energy flux $F_E$ across the dataset on a symmetric log scale. Values with $|F_E| < 10$~eV/s are binned to zero. The vertical dashed line marks the classification threshold at $10^5$~eV/s.}
    \label{fig:flux_hist}
\end{figure}

\begin{figure}[h]
    \centering
    \includegraphics[width=0.7\linewidth]{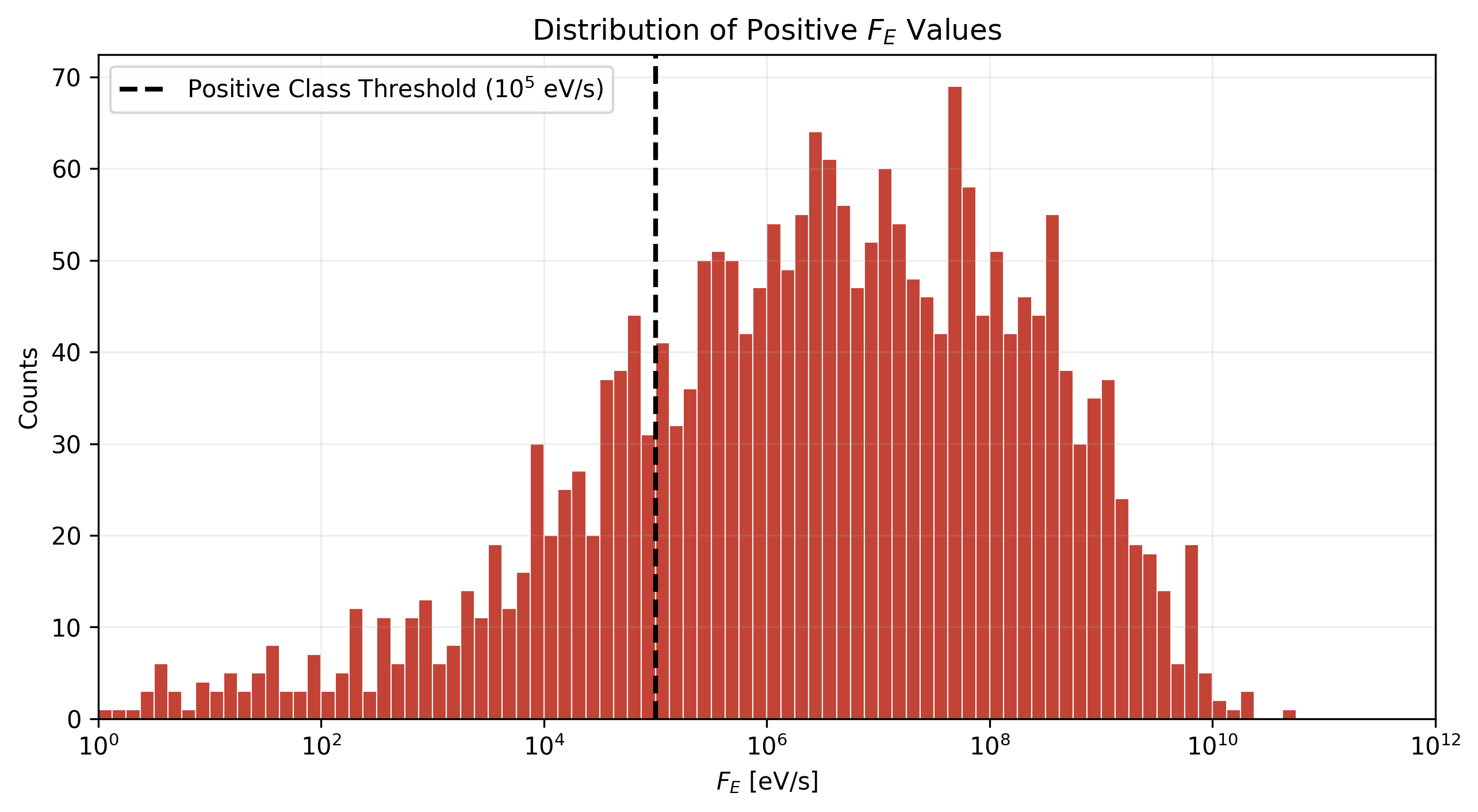}
    \caption{Close-up of the positive-flux region. The threshold at $10^5$~eV/s separates weak positives from configurations with substantial energy absorption.}
    \label{fig:flux_closeup}
\end{figure}

\Cref{fig:flux_hist} shows the distribution of energy flux $F_E$ across the dataset, plotted on a symmetric logarithmic scale (values with $|F_E| < 10$~eV/s are binned to zero). The distribution is strongly asymmetric: the vast majority of configurations are dissipative ($F_E < 0$), with magnitudes concentrated around $10^6$-$10^{10}$~eV/s, while positive-flux configurations are comparatively rare. \Cref{fig:flux_closeup} provides a close-up of the positive-flux region, showing that the bulk of energy-absorbing configurations produce fluxes in the range $10^3$-$10^{10}$~eV/s.

The threshold $F_E > 10^5$~eV/s was chosen as a conservative lower bound for physically meaningful energy absorption rates.
Systems exhibiting positive bosonic flux of practical interest typically produce values in the range $10^7$-$10^{10}$~eV/s, several orders of magnitude above the threshold.
Near-zero fluxes ($|F_E| \ll 10^5$~eV/s) are ambiguous: they may arise from experimental noise in steady-state measurements.
By placing the threshold well below the regime of interest, we ensure that the positive fluxes are distinctively defined.

\FloatBarrier
\section{Hyperparameter Optimization}
\label{app:hyperparams}

Hyperparameters were optimized using the Optuna framework~\cite{optuna_2019}, which employs a Tree-structured Parzen Estimator (TPE) to efficiently navigate the search space.
The objective was PR-AUC on the validation set, with models trained on the 28{,}000-graph training pool. 
\Cref{tab:hp_search} gives the search ranges, the selected values, and the settings held fixed.

\begin{table}[h]
    \centering
    \caption{Search space of the primary hyperparameter search, its selected values, and the settings held fixed. All continuous parameters were drawn on a log scale.}
    \label{tab:hp_search}
    \begin{tabular}{lcc}
        \toprule
        \textbf{Hyperparameter} & \textbf{Search Range} & \textbf{Selected} \\
        \midrule
        Latent dimension ($z$) & $\{32, 64, 128\}$ & 32 \\
        Hidden dimension ($H$) & $\{64, 128, 256\}$ & 128 \\
        Learning rate & $[10^{-4},\; 5\times10^{-3}]$ & $9.48 \times 10^{-4}$ \\
        Training batch size & $\{32, 64, 128\}$ & 128 \\
        $\beta_\text{recon}$ & $[1.0,\; 200.0]$ & 2.674 \\
        $\beta_\text{KL}$ & $[10^{-4},\; 10^{-1}]$ & $5.90 \times 10^{-3}$ \\
        $\beta_\text{pred}$ & $[1.0,\; 300.0]$ & 65.84 \\
        \midrule
        \multicolumn{3}{@{}l@{}}{\emph{Fixed, not searched}} \\
        MC-dropout rate & --- & 0.2 \\
        GAT message-passing layers & --- & 2 \\
        Attention heads (layer 1 / output) & --- & 2 / 1 \\
        Graph pooling & --- & attention \\
        Stage-0 pre-train epochs & --- & 30 \\
        \bottomrule
    \end{tabular}
\end{table}

Training used the AdamW optimizer~\cite{loshchilov2019decoupled} with a \texttt{ReduceLROnPlateau} scheduler on the training loss (patience 10 epochs, factor 0.5).
Final models are trained with five random seeds, each with early stopping on validation PR-AUC (patience 40, maximum 400 epochs); across those seeds the selected configuration converged after 263 epochs on average (range 133 - 341).
Search trials were capped at 300 epochs and two hours; 24 completed and 105 were pruned, for 38.8 GPU-hours.
The two-hour cap was binding for batch-size-32 trials, which stopped at $\sim$128 epochs against 273 for batch-128.
The search was run on the GVAE and the resulting configuration transferred to the baselines, which were not separately tuned; for the two discriminative baselines the three loss weights are inert and the stage-0 autoencoder pre-train is skipped.

A second search over the positive-class weight, with the architecture frozen at
the selection above, found no configuration that improved validation PR-AUC
over the unweighted objective, which is reported throughout.

 \FloatBarrier
\section{Ablations}
\label{app:ablations}

Each cell changes one factor of the selected configuration and retrains five seeds.
Differences are paired on seed -- same initialization, same training rows, one hyperparameter changed -- which removes the between-seed variance that dominates a five-run comparison and reduces the standard error of a difference from roughly 0.03 to 0.01--0.03.
\Cref{tab:ablations} reports differences against the baseline on the interpolation test set; an asterisk marks a difference exceeding two standard errors.

\begin{table}[h]
    \centering
    \caption{Paired ablation differences on the interpolation test set, five seeds per cell. Parenthesised values are paired standard errors over those five seeds; the baseline figures carry sample standard deviations (PR-AUC $0.4089 \pm 0.0399$, ROC-AUC $0.9204 \pm 0.0036$). Cells are GVAE unless stated.}
    \label{tab:ablations}
    \begin{tabular}{lcc}
        \toprule
        \textbf{Change} & \textbf{$\Delta$ PR-AUC} & \textbf{$\Delta$ ROC-AUC} \\
        \midrule
        $\beta_\text{KL} = 0$ & $-0.017\ (0.010)$ & $+0.000\ (0.003)$ \\
        $\beta_\text{KL} \times 10$ & $-0.108\ (0.037)^*$ & $-0.023\ (0.010)^*$ \\
        $\beta_\text{recon} / 10$ & $-0.083\ (0.017)^*$ & $-0.022\ (0.011)^*$ \\
        Latent dimension 64 & $-0.046\ (0.023)$ & $-0.004\ (0.007)$ \\
        Latent dimension 128 & $-0.138\ (0.028)^*$ & $-0.027\ (0.007)^*$ \\
        Latent dimension 64, GAT & $+0.038\ (0.058)$ & $+0.028\ (0.025)$ \\
        \bottomrule
    \end{tabular}
\end{table}

The reconstruction term is load-bearing: weakening it tenfold costs $0.083$ PR-AUC.
The variational term is not, at least for prediction: removing it entirely is within noise on the interpolation and extrapolation test sets - the only difference clearing two standard errors is $-0.003\,(0.001)$ ROC-AUC on extrapolation, detectable but negligible - while raising it tenfold is severely damaging, so the selected value sits on a plateau on one side and a cliff on the other.

The KL term does, however, organize the latent space, which is the architecture's stated purpose.
Comparing $\beta_\text{KL} = 0$ against the baseline over five seeds, what the latent \emph{encodes} is unchanged - temperature is linearly decodable at $R^2 = 0.993 \pm 0.004$ against $0.995 \pm 0.002$ - but how it is \emph{arranged} degrades: the variance captured by the first principal component falls from $0.482 \pm 0.082$ to $0.372 \pm 0.056$, and temperature's rank correlation with a leading principal component falls from $0.984 \pm 0.014$ to $0.928 \pm 0.039$ (3.8 standard errors).
The arrangement also becomes seed-dependent: $\beta_0$'s correlation with a leading component spans $0.08$ - $0.54$ across the five unregularized seeds against $0.40$ - $0.73$ with regularization, so whether a projection displays a $\beta_0$ gradient would depend on the seed chosen.

Latent dimension 32, selected by the search in all fifteen top trials, is confirmed for the GVAE: 128 is worse on every split and 64 is directionally worse within noise.
For the discriminative GAT the latent dimension does not matter, with 64 marginally positive and all cells within noise, so this is a property of the generative model rather than of the task.

Finally, periodically replacing an acquisition with random selection - every eleventh cycle - leaves the uncertainty acquisitions unaffected: restricted to those cycles the per-cycle positive rate is $0.280 \pm 0.005$ without injection against $0.281 \pm 0.008$ with it, a difference of $0.2$ standard errors, and pairing on cycle index agrees (injection higher in 17 of 30 indices, sign test $p = 0.59$).
The design detects relative differences above 4.7\%.
Its only effect on a trajectory is the direct cost of the cycles spent at the base rate, two of 32 returning 3.4 - 4.2\% positives against 25 - 36\% for uncertainty cycles.

\FloatBarrier
\section{Baseline Comparison}
\label{app:baselines}

\Cref{tab:baselines} compares the GVAE against three alternatives on the natural dataset: a
plain graph autoencoder (AE) with the same encoder and decoders but no
variational term, and two purely discriminative models (GAT and GCN
classifiers) with no decoders at all. All four use the configuration selected
for the GVAE (\cref{app:hyperparams}) and are scored on the same held-out sets.

\begin{table}[h]
    \centering
    \caption{Predictive performance on the natural dataset, five seeds
    per model. $\pm$ is the sample standard deviation over seeds.}
    \label{tab:baselines}
    \small
    \begin{tabular}{lcccc}
        \toprule
         & \multicolumn{2}{c}{\textbf{Interpolation}} & \multicolumn{2}{c}{\textbf{Extrapolation}} \\
        \cmidrule(lr){2-3}\cmidrule(lr){4-5}
        \textbf{Model} & ROC-AUC & PR-AUC & ROC-AUC & PR-AUC \\
        \midrule
        GVAE            & $0.920 \pm 0.004$ & $0.409 \pm 0.040$ & $0.895 \pm 0.010$ & $0.339 \pm 0.031$ \\
        AE              & $0.915 \pm 0.006$ & $0.354 \pm 0.014$ & $0.890 \pm 0.010$ & $0.312 \pm 0.015$ \\
        GAT classifier  & $0.876 \pm 0.065$ & $0.287 \pm 0.104$ & $0.862 \pm 0.041$ & $0.255 \pm 0.065$ \\
        GCN classifier  & $0.896 \pm 0.004$ & $0.286 \pm 0.009$ & $0.879 \pm 0.004$ & $0.257 \pm 0.007$ \\
        \bottomrule
    \end{tabular}
\end{table}

Two patterns separate the models, and both appear only in PR-AUC. The two
architectures carrying auxiliary decoders outperform the two purely
discriminative ones by $0.07$-$0.12$ PR-AUC on the interpolation test set,
consistent with \cref{app:ablations}, where weakening the reconstruction term
tenfold costs $0.083$. Among the two generative models the variational term
adds a further $0.055$ (s.e.\ $0.019$), though on extrapolation that narrows to
$0.027$ (s.e.\ $0.015$) and is no longer resolvable. ROC-AUC barely separates
the GVAE and the AE ($0.920$ against $0.915$), which is expected at a 
$4.2\%$ base rate: PR-AUC is the metric sensitive to performance on the rare
class. The GAT classifier is also by far the least stable model, with a seed
spread of $\pm 0.104$ test PR-AUC against $\pm 0.040$ for the GVAE and
$\pm 0.009$ for the GCN, so its point estimate should be read with that in mind.

The baselines were not separately tuned: the configuration of
\cref{app:hyperparams} was selected on GVAE validation PR-AUC and transferred,
and for the two discriminative models the three loss weights are inert and the
stage-0 pre-train is skipped. The ablation bounds the size of that handicap --
giving the GAT classifier its preferred latent dimension of $64$ recovers
$+0.038$ (s.e.\ $0.058$) PR-AUC, which does not close a $0.12$ gap. The GVAE is
therefore not merely competitive on prediction; it is also the only model here
that yields the structured latent space of \cref{fig:latent} and the MC-dropout
uncertainty the acquisition loop runs on.
\FloatBarrier
\section{Enrichment by Topology}
\label{app:topology_enrichment}

The aggregate enrichment reported in \cref{sec:results} averages over eight
device geometries whose positive base rates differ by nearly a factor of four.
\Cref{tab:topology_enrichment} separates them.

\begin{table}[h]
    \centering
    \caption{Hit rate within each topology, pooled over three trajectories per
    method and over all cycles; rates are computed over each strategy's own
    acquisitions. The random column is the random strategy's hit rate on its
    own acquisitions, which differs slightly from the independent sample of
    \Cref{tab:scarcity}. Budget share is the fraction of all acquisitions the
    active-learning loop placed in that topology; the normalized Shannon
    entropy of those shares is $0.948$ against $1.000$ for the uniform control,
    and GP-BO places 12.5\% in each topology by construction. Bold marks the
    highest hit rate among the three search methods in each row. No per-cell
    uncertainty is quoted because the rates are pooled rather than averaged
    over trajectories.}
    \label{tab:topology_enrichment}
    \footnotesize
    \setlength{\tabcolsep}{4pt}
    \begin{tabular}{@{}lcccc ccc c@{}}
        \toprule
        & \multicolumn{4}{c}{\textbf{Hit rate}} & \multicolumn{3}{c}{\textbf{Enrichment}} & \textbf{AL} \\
        \cmidrule(lr){2-5}\cmidrule(lr){6-8}
        \textbf{Topology} & Random & EI & UCB & AL & EI & UCB & AL & \textbf{budget} \\
        \midrule
        $1{\times}2$ & 0.044 & 0.060 & 0.097 & \textbf{0.309} & $1.35\times$ & $2.21\times$ & $6.99\times$ & 17.3\% \\
        $1{\times}3$ & 0.064 & 0.294 & \textbf{0.414} & 0.335 & $4.63\times$ & $6.50\times$ & $5.26\times$ & 18.0\% \\
        $1{\times}4$ & 0.078 & \textbf{0.434} & 0.397 & 0.327 & $5.57\times$ & $5.09\times$ & $4.20\times$ & 23.9\% \\
        $2{\times}2$ & 0.038 & 0.133 & 0.097 & \textbf{0.273} & $3.51\times$ & $2.55\times$ & $7.19\times$ &  9.0\% \\
        $2{\times}3$ & 0.033 & 0.203 & 0.216 & \textbf{0.256} & $6.15\times$ & $6.54\times$ & $7.75\times$ &  7.8\% \\
        $3{\times}2$ & 0.022 & 0.110 & 0.068 & \textbf{0.175} & $4.94\times$ & $3.04\times$ & $7.89\times$ &  7.3\% \\
        $2{\times}4$ & 0.037 & 0.168 & 0.161 & \textbf{0.192} & $4.56\times$ & $4.38\times$ & $5.22\times$ & 10.5\% \\
        $3{\times}3$ & 0.020 & 0.059 & 0.086 & \textbf{0.171} & $2.91\times$ & $4.23\times$ & $8.35\times$ &  6.3\% \\
        \bottomrule
    \end{tabular}
\end{table}

Every method enriches over its random control in every topology. For the
active-learning strategy the factor ranges from $4.20\times$ to $8.35\times$;
the GAT-classifier strategy behaves the same way qualitatively, with its
lowest enrichment on $1{\times}4$ and its highest on a sparse grid
($3{\times}2$, at $11.11\times$).

Two features of the table are worth stating. First, the base rate is set by the
number of rows rather than by system size: single-row systems carry
0.044--0.078 positives per call, two-row systems 0.033--0.038, and three-row systems 0.020--0.022, 
with $1{\times}4$ and $2{\times}2$ differing by a factor
of two at equal $N$. Second, enrichment is smallest where the base rate is
already highest - $4.20\times$ at $1{\times}4$ against $8.35\times$ at $3{\times}3$ - 
so the method contributes most in the geometries where positives
are rarest.

The comparison with GP-BO is sharper per topology than in aggregate. The
uncertainty-driven strategy has the higher hit rate in seven of the eight
topologies against EI and six of eight against UCB; the exceptions are the
longest chains, $1{\times}4$ for both acquisitions and $1{\times}3$ for UCB,
which are also the two richest topologies. In the five geometries with more
than one row the active-learning rate exceeds the Gaussian-process rate by
factors of 1.1 to 2.9, and it does so while placing a \emph{smaller} share of
its budget there, 6.3-10.5\% against the baseline's uniform 12.5\%. The
Gaussian-process profile is also far less even, from $1.35\times$ on $1{\times}2$ to $6.54\times$ on $2{\times}3$, where the active-learning
strategy stays between $4.20\times$ and $8.35\times$ throughout.

The budget shares show the loop reallocating rather than sampling uniformly: it
places 23.9\% of its acquisitions in $1{\times}4$ and 6.3\% in $3{\times}3$.
That freedom is unavailable to a Gaussian process, which must divide its budget
evenly because the input dimension changes with $N$. It does not, however,
account for the aggregate gap: the loop spends more of its budget on
$1{\times}4$ and still loses there, and less on the 2D grids and still wins by
up to $2.9\times$.

\section{Computational Resources}
\label{app:resources}

All experiments were conducted on a high-performance computing cluster. The hardware and software environment included:

\begin{itemize}
    \item \textbf{Hardware:} A single node equipped with two \textbf{NVIDIA A100-SXM4 GPUs} with 40 GB VRAM each.
    \item \textbf{Software Stack:} The models were implemented in \textbf{Python 3.12} using \textbf{PyTorch} and \textbf{PyTorch Geometric}.
    \item \textbf{Model Complexity:} The GraphVAE architecture consists of \textbf{120,488 trainable parameters}, with an estimated model size of \textbf{0.48 MB}.
    Each AL cycle, which included the training of the model, took a median of 57 minutes.
    \item \textbf{Data Handling:} The pipeline processed a dataset of \textbf{40,000 graphs}, partitioned into a training set of \textbf{28,000 samples}, a validation set of \textbf{6,000 samples}, and an interpolation test set of \textbf{6,000 samples}.
    \item \textbf{Execution Environment:} Computational jobs were managed via the \textbf{SLURM} workload manager with auto-requeueing enabled to ensure job persistence across iterative Active Learning cycles.
\end{itemize}

\subsection{Oracle Cost and Scaling}
\label{app:oracle_cost}

To quantify the cost of the physics oracle, we measured the wall-clock time of a single master-equation evaluation as a function of system size (\cref{tab:oracle_timing}).
Each value is a median over repeated serial calls on CPU, with 8 cores allocated via SLURM. The $3{\times}4$ entry is the median of eleven independent single-topology generation runs totalling 4{,}850 oracle calls (per-call range 70.2-81.3~s).

\begin{table}[h]
    \centering
    \caption{Per-call wall-clock time of the master-equation oracle on CPU. $N$ is the number of orbitals and $2^N$ the dimension of the many-body Fock space; see the text for how each entry was measured.}
    \label{tab:oracle_timing}
    \begin{tabular}{lccc}
        \toprule
        System & $N$ & Fock dimension $2^N$ & Median time per call \\
        \midrule
        $1\times2$ & 2  & 4    & 0.77\,ms \\
        $1\times3$ & 3  & 8    & 1.5\,ms  \\
        $2\times2$ & 4  & 16   & 3.5\,ms  \\
        $2\times3$ & 6  & 64   & 32\,ms   \\
        $3\times3$ & 9  & 512  & 0.65\,s  \\
        $3\times4$ & 12 & 4096 & 76\,s    \\
        \bottomrule
    \end{tabular}
\end{table}

The per-call cost grows by a factor of roughly $1\times10^5$ between $N=2$ and $N=12$, and the growth steepens with system size.
The steady state, i.e.\ the null space of the $2^N\times2^N$ rate matrix, is obtained by a dense singular value decomposition whose cost scales as $(2^N)^3$; this step dominates the cost of a call for the largest systems, while for smaller systems the remaining steps of the calculation take most of the time.
This exponential growth is consistent with the scaling reported for related many-body transport calculations~\cite{iv2024emergence}.

Extrapolating the asymptotic cubic scaling from the measured $N=12$ value, a single evaluation would take approximately 10 minutes at $N=13$, 1.4 hours at $N=14$, and 3.6 days at $N=16$.
These are estimates rather than measurements, but they show that exhaustive or random screening of larger devices quickly becomes intractable.
This is the regime that motivates a surrogate-guided search, in which the expensive oracle is queried only for the candidates selected by the active-learning loop.


\end{document}